An interpretable framework using foundation models for fish sex identification

Zheng Miao, Tien-Chieh Hung*

Department of Biological and Agricultural Engineering, University of California Davis, 1 Shields Ave., Davis, CA 95616, USA

∗Corresponding Author: thung@ucdavis.edu

Abstract

Accurate sex identification in fish is vital for optimizing breeding and management strategies in aquaculture, particularly for species at the risk of extinction. However, most existing methods are invasive or stressful and may cause additional mortality, posing severe risks to threatened or endangered fish populations. To address these challenges, we propose FishProtoNet, a robust, non-invasive computer vision–based framework for sex identification of delta smelt (Hypomesus transpacificus), an endangered fish species native to California, across its full life cycle. Unlike the traditional deep learning methods, FishProtoNet provides interpretability through learned prototype representations while improving robustness by leveraging foundation models to reduce the influence of background noise. Specifically, the FishProtoNet framework consists of three key components: fish regions of interest (ROIs) extraction using visual foundation model, feature extraction from fish ROIs and fish sex identification based on an interpretable prototype network. FishProtoNet demonstrates strong performance in delta smelt sex identification during early spawning and post-spawning stages, achieving the accuracies of 74.40% and 81.16% and corresponding F1 scores of 74.27% and 79.43% respectively. In contrast, delta smelt sex identification at the subadult stage remains challenging for current computer vision methods, likely due to less pronounced morphological differences in immature fish. The source code of FishProtoNet is publicly available at: https://github.com/zhengmiao1/Fish_sex_identification.

1. Introduction

Delta smelt (*Hypomesus transpacificus*) is a small, silvery fish native to the San Francisco Bay-Delta estuary in California. The delta smelt plays a vital role as an indicator species for the health of the delta ecosystem, owing to its high sensitivity to alterations in water quality and habitat conditions. However, delta smelt has been listed as a critically endangered species by the state of California (CNDDB 2024), as its population has severely declined due to habitat loss, water diversions and pollution. The long-term recovery of the delta smelt population in the wild remains challenging, given the short lifespan and low natural reproductive success of delta smelt (Bennett et al., 2005). Consequently, artificial aquaculture of delta smelt plays a critical role in population stabilization and growth of this endangered fish (Lindberg et al., 2013). To ensure the population growth of fish, accurate sex identification is vital important for fish optimizing both reproduction and propagation in hatcheries (Budd et al., 2015). In this context, a convenient way to accurate sex determination is essential to achieve sex ratios control and successful reproduction of fish, thereby maintaining genetic diversity and improving breeding efficiency (Wang et al., 2025; Lema et al., 2024). Determining fish sex is also essential for understanding sex-based differences in immune responses, which can improve disease management, welfare, and treatment strategies in aquaculture (Caballero-Huertas et al., 2025).

Several approaches have been developed to identify the sex of fish based on the morphological characteristics of their reproductive organs. Gonda biopsy or dissection can be used to distinguish the female and male of fish by the examination of fish gonads (Mansur et al., 2021). Endoscopy, a tube-like optical device equipped with a camera and light source, was used to visually inspect gonads internally (Swenson et al., 2007). Endo et al. (2024) used ultrasonography to determine the sex of young carp (*Cyprinus carpio*). However, these methods still rely on experts to further distinguish the testis and ovary of the fish. At the same time, fish sex determination methods based on reproductive organ are invasive, potentially causing physical stress to the fish.

Advancement and development of computer vision techniques and artificial intelligence (AI) offers effective tools for fish sex recognition by morphological differences. Several studies have employed machine learning (ML) methods such as support vector machines (SVMs) and deep neural networks for sex classification based on phenotype in aquatic species. Hosseini et al. (2019) applied SVMs to differentiate the sex of zebrafish (*Danio rerio*) based on caudal fin coloration features. They also employed deep convolutional neural networks (DCNNs) to extract visual feature from the fish images to recognize the sex of zebrafish. A multilayer perceptron (MLP) trained by de Cerqueira & de Santana (2023) achieved over 80% precision, recall, F1-score, and accuracy in classifying fish sex of a commercial fish species in Bazil. To determine sex of Chinese mitten crabs (*Eriocheir sinensis*), Chen et al. (2023) proposed a lightweight classification and detection architecture, GMNet-YOLOv4. Their model replaced the standard convolutional layers with depth-wise separable convolutions from MobileNet, achieving an average precision of 97.68% for male crab detection and 97.23% for female crab detection.

In practice, it remains challenging for machine learning to precisely identify the fish sex when there are no obvious morphological differences between female and male of fish can be observed by humans, as in the case of delta smelt. Two critical issues are particularly important in fish sex identification based on machine learning methods. The first is to how to effectively extract features or representations from the images. The second is how to make accurate and reliable decisions for fish sex identification based on these representations. To tackle these issues, we introduce corresponding solutions in the subsequent sections.

Visual fish sex identification methods based on machine learning are easily affected by the background noises, which prone to memorize the bias of different backgrounds while overlooking the intrinsic difference of two fish sex, even though they may still achieve good prediction performance on specific fish datasets. Therefore, we only want to extract the effective feature from fish itself to determine the fish sex.

To mitigate interference from background noise, it is essential for fish regions of interests (ROIs) are extracted. Object detection methods or instance segmentation methods can be used to generate proposal fish regions by bounding boxes or segmented masks. One-stage methods, like YOLO (Jiang et al., 2022), and two-stage methods (e.g. Faster RCNN, Ren et al 2016) are two common architectures to be used for object detection methods. However, these close-set detection and segmentation methods are trained on specific species fish, which is costly for data collection and data annotation for each fish species. In contrast, open-vocabulary visual foundation models, such as Grounding DINO and Segment Anything Model (SAM), enable the zero-shot open-set recognition via visual prompts or text prompts. Grounding DINO (Liu et al., 2024) is an open vocabulary object detection method by incorporating textual descriptions of objects. Grounding

DINO enables open-set object detection by incorporating natural language to identify unseen objects, in contrast to traditional closed-set object detectors. Segment Anything Model (SAM, Kirillov et al., 2023), a visual foundation model developed by Meta, introduces the prompt-based paradigm to the segmentation task, overcoming the limitations of traditional segmentation methods on class-specific categories. SAM was trained on over 1 billion masks across 11 million images, exhibiting strong generalization capabilities for downstream tasks (Kirillov et al., 2023). Segment Anything Model 2 (SAM2) (Ravi et al., 2024) extends the capabilities of the original SAM from static image segmentation to video-based segmentation, achieving an inference rate of 64.8 (frames per seconds) FPS. SAM2 is capable of handling complex backgrounds and diverse objects due to its enhanced feature extraction and multimodal fusion capabilities. In this study, Grounding DINO and SAM2 are combined to zeros-shot generate the ROIs based on text prompts，in order to alleviate the bias from background.

The other critical limitation of existing deep learning methods for fish sex recognition is the deficiency in model interpretability of fish sex determination, which poses a significant obstacle for biologists and marine scientists. To address this issue, some post-hoc explanation methods have been proposed to interpret classification decision by highlighting class-discriminative regions within image, such as class activation mapping (CAM, Zhou et al., 2016) and gradient-weighted class activation mapping (Grad-CAM, Selvaraju et al., 2017). However, these post-hoc explanation methods often exhibit low faithfulness (Jesus et al., 2021; Bordt et al., 2022) in model explanation. Specifically, the regions highlighted by these post-hoc methods do not align with the actual regions responsible for the model prediction, thereby increasing the risk of potential misinterpretation (Krishna et al., 2022).

Correspondingly, prototype learning provides an effective way to explain the decision-making process for categories classification by introducing the phenotype-based learning (Yang et al., 2018). Prototype learning methods have been applied fine-grained object recognition in zero-shot or few-shot classification (Ji et al., 2023; Liu et al., 2023) due to its capacity to distinguish subtle difference between similar categories and perform well even in few labeled samples in dataset. Prototype-based learning is paradigm that first obtains representative samples (prototypes) for each class and then makes predictions by comparing distance between new samples and learned prototype representations in metric space (Snell et al., 2017). A decision tree model combining neural network was proposed by Nauta et al., (2021) to significantly reduce the number of phenotypes learned in phenotype-learning.

In this work, we explored the potential of computer-assisted techniques to identify the sex of fish, using delta smelt as a case study. We hypothesize that morphological differences between female and male delta smelt can be used as distinguishing patterns for sex determination, even it is hard to be distinguished visually by human. An explainable AI model FishProtoNet was developed in this study to use the tiny morphological differences between female and male across various life stages to achieve sex identification of delta smelt. The main contributions of this study are as follows:

1) Visual foundation models were used to generate category-agnostic fish ROI proposals in a zero-shot manner by combining Grounding DINO with SAM2, effectively mitigating interference from background noise.

2) An interpretable prototype-based learning network was developed to enhance transparency in the decision-making process for fish sex identification.

3) Systematic data augmentation strategies were systematically selected to improve model robustness in fish sex identification.

2. Methodology

2.1 Model Development

We designed a model architecture of FishProtoNet for delta smelt sex identification, as illustrated in Figure 2, which is mainly constructed of three modules: (1) an ROI extraction module, (2) a feature extraction module, and (3) a sex prediction module. Data augmentation is adopted between the ROI extraction module and the feature extraction module. To accurately localize the ROIs, we first adopted Grounding DINO to perform zero-shot detection of delta smelt. The Grounding DINO was first used to generate the bounding boxes around fish with the text prompt "fish". This module leverages the representational power of a foundation model pre-trained on a large-scale image-text corpus. Then SAM2 was employed to obtain the fish masks with the visual prompt of object detection bounding boxes from Grounding DINO. The obtained fish masks will be used as the ROIs of delta smelt for subsequent processing, thus effectively reducing the background interference. Data augmentation is performed on proposed ROIs to increase the variance of training data samples. In the feature extraction neck, we employed the backbones of ResNet, excluding their fully connected layers, to extract high-dimensional semantic features from the fish masks generated in the ROI extraction stage, enabling effective representation of fish-related information. In the classifier head, an explainable AI model combining deep neuron decision tree with prototype-based methods was trained to identify the fish sex using the feature extracted from the neck stage. The soft neural decision tree based on prototype learning provided an explanation of the decision-making process for sex determination in delta smelt. Thus, a viable scheme for delta smelt sex identification was proposed using a prototype-based decision-making approach. The details of each module in our system are described in the following sections 2.1.1-2.1.4.

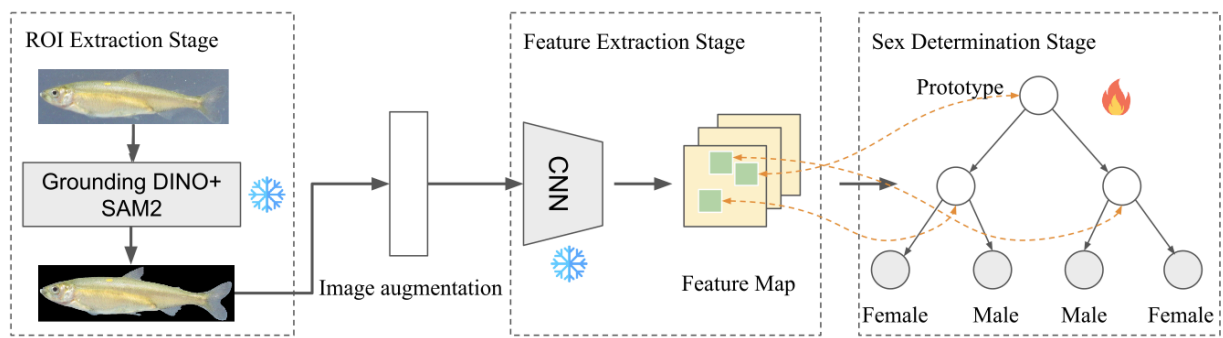

Figure 1. Framework of the proposed sex identification model. (Notes: snowflake icon means no weight updates of the neural network and fires icon means the weights updates on this part of neural network)

2.1.1 ROI Extraction via Foundation Models

Accurate fish ROIs extraction provides informative feature patterns for accurate fish sex identification while reducing interference from background noise. In the ROI extraction stage, we first use grounding DINO to obtain the fish bounding boxes locations with the text prompt, then we obtain the fish mask region using the SAM2 method based on these bounding boxes, as the figure 2 illustrated.

Specifically, we use pre-trained grounding DINO with the "fish" text prompt. Swin Transformer was adopted as image encoder to extract the image feature while BERT was used as text encoder to obtain text embedding features. The extracted image and text features were enhanced to text-aware visual embeddings and image-aware text embeddings through feature enhancer modules consist of multiple self-attention layers. The text embeddings can be used to guide the visual feature relevant to "fish". Next, the fish bounding boxes were predicted by the decoder module at the end of grounding DINO.

Then we used predicted fish bounding boxes as the visual prompts of SAM2 to get the fish ROIs as figure 3 shown. Following to (Ravi et al., 2024), we employed a pre-trained vision transformer (ViT) model based on masked autoencoders (MAE) to extract image features. The fish bounding boxes were feed into the mask decoder, which decodes the image feature to generate fish ROIs.

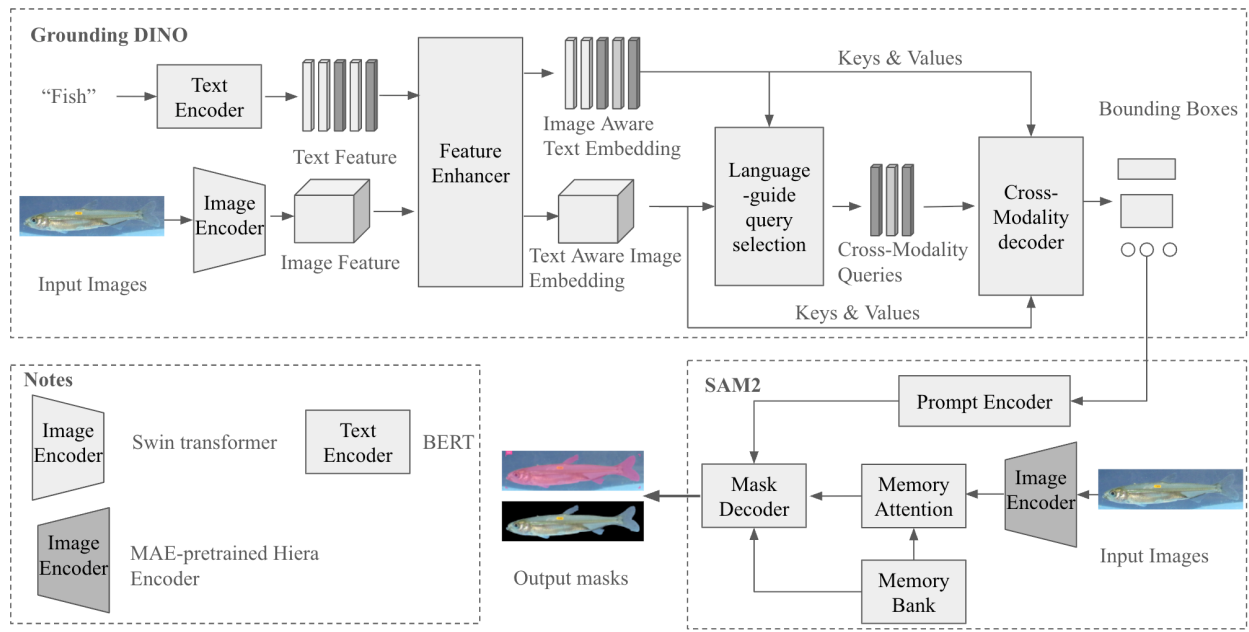

Figure 2. Grounding DINO-guided SAM2 for delta smelt proposal generation (ROIs of delta smelt generations). The module consists of Grounding DINO and SAM2. Grounding DINO was used to predict the object bounding box using the text prompt "fish." The predicted bounding box was then passed to the SAM2 model as a visual prompt to generate a segmentation mask of the delta smelt.

2.1.2 ROI-level Data Augmentation

Models with large number of parameters in machine learning are prone to overfitting when trained on small datasets. Appropriate data augmentation can mitigate memorization of irrelevant noise and encourage the model to learn meaningful and discriminative features for fish sex identification.

Therefore, we applied a series of data augmentation techniques to the fish ROIs following the ROI extraction stage. The purpose of data augmentation is to increase dataset diversity and thereby mitigates the risk of model overfitting when training on the relatively small dataset. As the baseline augmentation, we adopted image resizing, normalization, and padding. On top of these, we conducted trials using additional augmentation operations, including random rotation, random horizontal flipping, color jitter, random cropping, and random erasing (Zhong et al., 2020). Each of them was individually combined with the base. Random rotation (±15 degrees) and horizontal flipping were used to enhance the model's robustness to variations in fish orientation and posture. Random cropping helped the model focus on local discriminative regions. Random erasing (Random erasing et al., 2020) was applied to address cases of missing visual information like the incomplete regions in the masks predicted during the ROI stage or to obscure specific visual elements, like visible implant alphanumeric (VIA) tags of the individual identity (Sandford et al., 2020) by randomly masking out rectangular regions of the image. These augmentations were applied during training to increase sample diversity across epochs and enhance model generalization. We also implemented color jitter on the base image, introducing perturbations such as changes in brightness, contrast, and saturation. Color jitter may help the model robust to the light changing in the practical aquaculture environment.

2.1.3 Feature Extraction

This study assumes that female and male delta smelt exhibit morphological and other phenotypical differences across developmental stages, arising from gonadal development and manifesting in body shape, coloration and texture. Consequently, efficient extraction of image features that capture these characteristics is critical for fish sex identification. Representation learning using deep neural networks has demonstrated enhanced feature extraction capabilities compared to traditional hand-crafted descriptors. In this study, we adopted a ResNet model pre-trained on ImageNet (He et al., 2016) to extract informative feature from fish ROIs related to these characteristics. The features extracted from the last convolutional block were used as input to the subsequent 2.1.4 sex prediction module. Based on transfer learning, we also fine-tuned the final fully connected layer to predict the fish sex in the comparison experiments.

2.1.4 Prototype-based Sex Prediction

Based on the feature extraction from fish ROIs, the sex prediction module is responsible for predicting the sex of delta smelt as figure 1 shown. This module can be implemented either as a single fully connected layer or a multilayer perceptron (MLP). A single fully connected layer performs well when the features are linearly separable, whereas an MLP is better suited for classifying nonlinearly separable feature classification. However, the decision-making process of these classifiers still lacks interpretability and transparency for fish sex identification due to their "black box" characteristics. Therefore, understanding how model makes fish sex predictions is particularly critical for delta smelt sex identification. Such interpretability enables domain experts, including biologists, oceanographers, aquaculture practitioners, to better understand and validate the model prediction.

To further enhance the transparency of model decision-making, we employed prototype learning methods, which are commonly used in few-shot learning for fine-grained classification. Prototype learning improves model robustness and interpretability by classifying new samples based on the similarity between their projected features and the learned class prototypes in metric space, offering an intuitive explanation for the predictions (Yang et al, 2018). Therefore, we trained a deep neural decision tree based on prototype learning (FishProtoNet) to predict the sex of delta smelt, following the train strategies in ProtoTree introduced by Nauta et al., (2021). The deep neural decision tree can significantly reduce the model parameters compared to fully connected layers. In the training process, each internal node in the neural tree encodes a prototype relate to fish sex identification. Each internal node of the neural tree network learns a visual prototype by computing the similarity between a learned prototype and embedded feature vectors extracted from fish images, as Formula (5) shown. Euclidean distance was employed as the metrics of similarities measurement. The leaf nodes in the tree represent fish sex probability distributions and reflect the aggregated information across the decision paths. The FishProtoNet model captures implicit relationships between the morphological features of different fish body regions and their corresponding sex labels. The Fish sex prediction module is a soft decision tree, predicting the final fish sex identity by aggregating the weighted probabilities of all the decision paths as the Formula (2) shown. Ultimately, fish sex is predicted based on the similarity between the test fish images samples and prototypes learned, considering all the decision paths. The cross-entropy loss function is used for the tree model parameter optimization as Formula (1) denotes.

$$\mathcal{L} = -\sum_{i=1}^{N}\sum_{c=1}^{C} y \log \hat{y}, \quad y \in \{0,1\} \tag{1}$$

Where L denote loss, y represents the ground-truth fish sex identity, and $\hat{y}$ denotes the predicted fish sex identify by models.

$$\hat{y}(x) = \sum_{n \in \mathcal{P}} \sigma(c_n)\, \pi_n(f(x; W)) \tag{2}$$

$$\pi_n(z) = \prod_{n \in \mathcal{P}} p_n(z) \tag{3}$$

$$p_n(z) = \exp(-\|\tilde{z}^* - p_n\|) \tag{4}$$

$$z^* = \arg \min_{(m,n) \in \mathcal{F}_i} \|z_{m,n} - p_k\| \tag{5}$$

Where x denotes the input images; n indexes the routing path number in the tree model; $\pi_n$ is the path in the tree model; W represents the parameters of feature extraction module; c denotes the leaf nodes output. $z^*$ denotes the learned prototypes, m, n index spatial locations on the feature map; c denotes the output of the leaf nodes, $\mathcal{F}_i$ denotes the corresponding feature map.

2.2 Evaluation metrics

We evaluated the model performance using multiple different metrics, including accuracy, precision, recall, and F1 score. The equations used to assess the performance of delta smelt sex prediction are listed below:

$$\text{Accuracy} = \frac{TP + TN}{TP + TN + FP + FN} \tag{6}$$

$$\text{Precision} = \frac{TP}{TP + FP} \tag{7}$$

$$\text{Recall} = \frac{TP}{TP + FN} \qquad (8)$$

$$\text{F1 Score} = 2 \cdot \frac{Precision * Recall}{Precision + Recall} \qquad (9)$$

where $TP$ (true positive) denotes the number of samples correctly predicted as belonging to the positive sex class, $TN$ (true negative) represents the number of samples correctly identified as belonging to the negative sex class, $FP$ (false positive) refers to the number of samples incorrectly classified as positive sex, and $FN$ (false negative) corresponds to the number of samples incorrectly classified as negative sex. Accuracy refers to the number of model correct prediction over the total number of samples in dataset. Precision is the ratio of true positive prediction over all the positive prediction. Recall (also called sensitivity) is the ratios of true positive prediction over all the actual positives. F1 score is the harmonic mean of precision and recall, representing a trade-off between precision and recall.

3. Experiments

3.1 Dataset

The images data of full life stages of delta smelt collected by Castillo et al., 2018 at the Fish Conservation and Culture Laboratory (FCCL; Byron, California), affiliated with the University of California, Davis. These data were collected over a four-month period during the winter–spring of 2013. The fish images were captured by using a Canon EOS Rebel XTi camera fitted with a 100-mm f/2.8 USM macro lens. These included partial side views of the body, which clearly display the visible implant alphanumeric (VIA) tags of the individual identity (Sandford et al., 2020). The delta smelt image dataset covers the development stage from subadult to adult, and it divided into three sessions: Session 1 represents the subadult stage prior to the spawning season, Session 2 corresponds to the early spawning season, and Session 3 points to the late or post-spawning stages. The delta smelt sex dataset contain 1603 images with 346 images in Session 1 (143 female and 203 male), 625 images in Session 2 (292 female and 333 male), and 632 images in Session 3 (300 female and 332 male). Some typical sample images in all sessions are shown in Figure 3. The FCCL delta smelt sex dataset was created by randomly partitioning the data into training and test sets at an 8:2 ratio for each developmental session, while preserving the original sex ratio in both subsets.

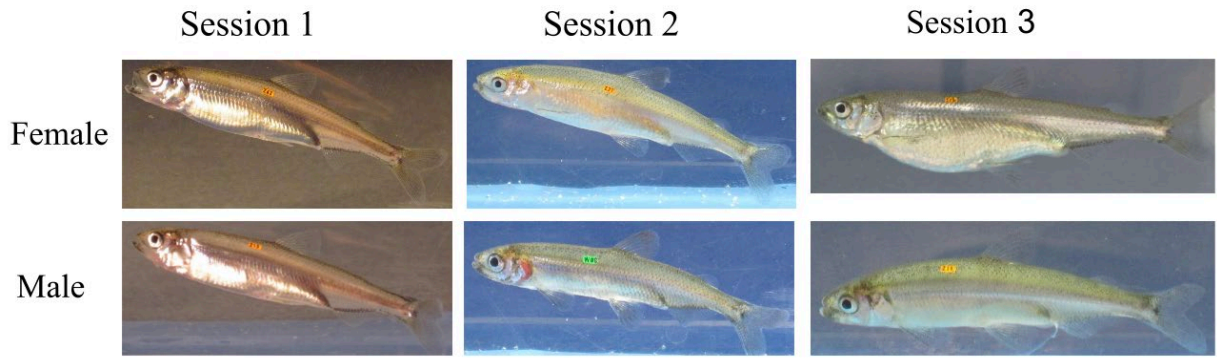

Figure 3. Image samples of delta smelt (Hypomesus transpacificus) across developmental and reproductive stages in the FCCL delta smelt sex dataset. Columns denote samples from different experimental sessions, illustrating the progression of sexual dimorphism throughout the life cycle : Session 1 (subadult), Session 2 (early spawning), and Session 3 (post-spawning). Rows distinguish between female and male specimens.

3.2 Model Training

The proposed explainable AI model was trained on the FCCL delta smelt sex dataset based on K-fold cross-validation, fully taking use of the data in the dataset. The K set to 5 in K-fold cross-validation experiments. To better distinguish delta smelt sex across different life stages, the model are trained separately across the different life stages.

At the ROI extraction stage, the pre-trained SAM2 with grounding DINO were used in a zero-shot manner to generate fish masks of delta smelt based on corresponding text prompts. That indicates the weights in the ROI extraction module requires no additional training, significantly reducing the computational costs of model training. Specifically, delta smelt bounding boxes were first predicted using the simple text prompts "fish". These fish bounding boxes were then used as visual prompts for SAM2 to predict fish masks from the original images. Fish ROIs were obtained by combining the predicted mask regions with the original images, as shown in Figure 4. ROI extraction was performed on each batch of image of fish dataset prior to data augmentation during data loading process.

Data augmentation was then used on the fish ROIs data during the training process to increase the data variability and alleviate potential overfitting. To keep the original ratios of the fish, we just add black padding to keep the ratios of original figure and input to the neural network. In details, data augmentation operations were progressively introduced to assess their effects on delta smelt sex classification performance, including random rotation, random crop and random erasing, as shown in Figure 4. The data augmentation operations were applied to each data batch during loading, prior to training the prototype neural tree for sex prediction, as illustrated in Figure 5.

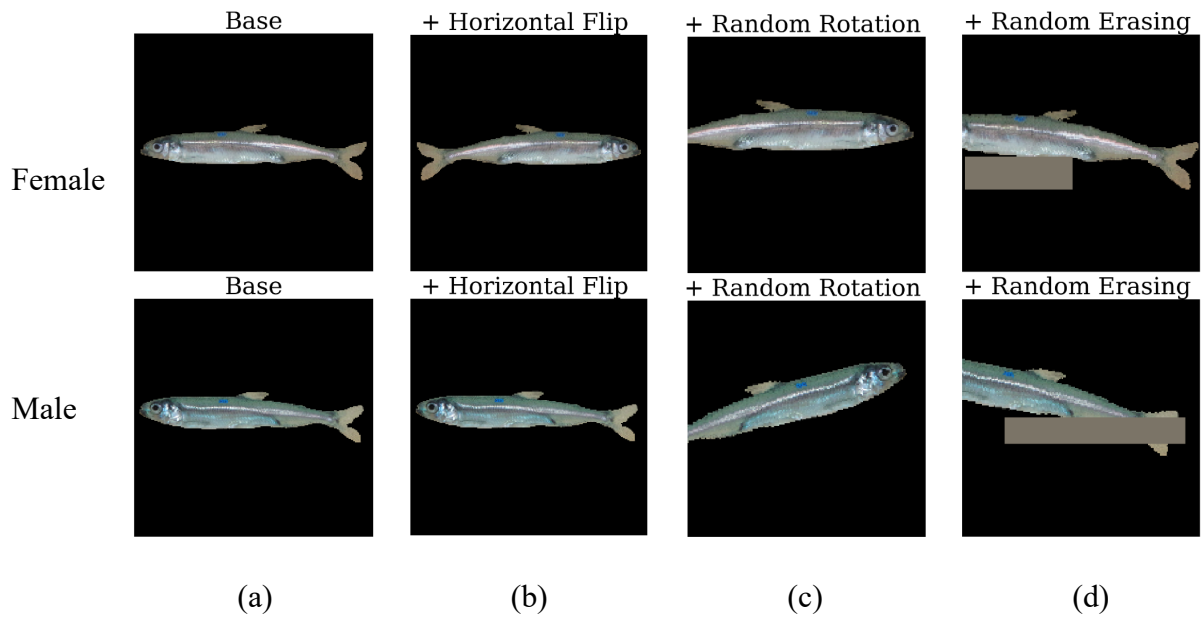

Figure 4. Visualization of data augmentation based on ROI extraction. (a) Base operation; (b) base operation with horizontal flipping; (c) random rotation applied on top of (b); and (d) random erasing applied on top of (c).

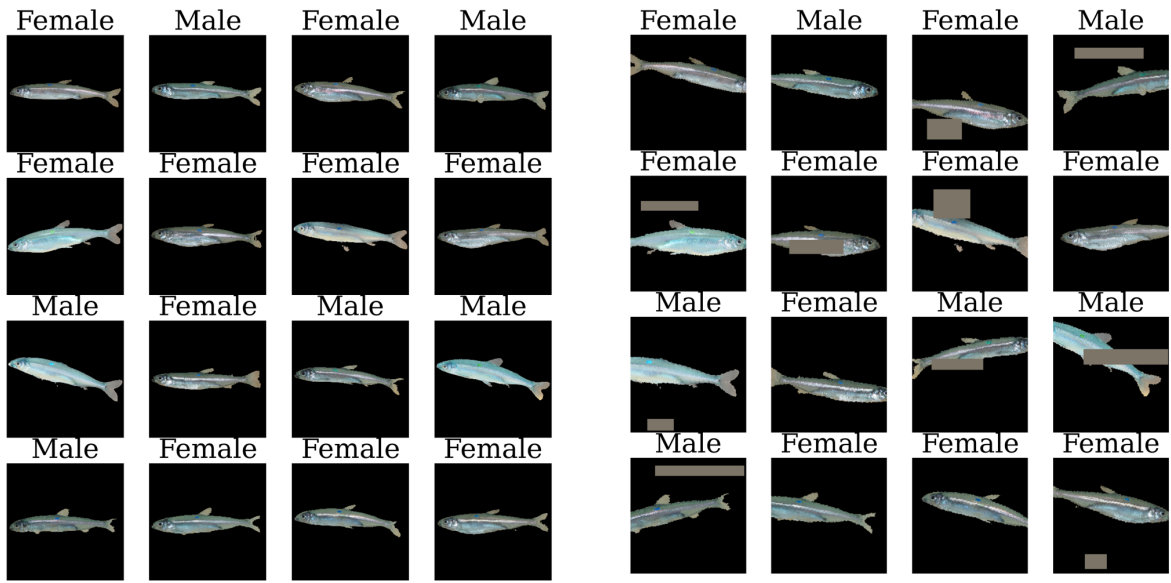

(a) Base batch  (b) Best data augmentation batch

Figure 5. Visualization of delta smelt data in a training batch: (a) fish ROIs with base data augmentation; (b) fish ROIs after ROI extraction and best data augmentation.

After data augmentation, we used a pre-trained ResNet-50 convolutional backbone to extract high-level semantic features from fish ROIs. Subsequently, a neural prototype tree model was trained as an interpretable classification network to identify delta smelt sex. FishProtoNet was trained independently across three sessions of the FCCL delta smelt sex dataset as illustrated in Figure 7, covering the life stages from subadult to adult Delta Smelt. Cross-entropy loss was used as loss function for model parameters optimization. The Adam optimizer was used with an initial learning rate of 0.001, OneCycle learning rate schedule, and batch size of 16.

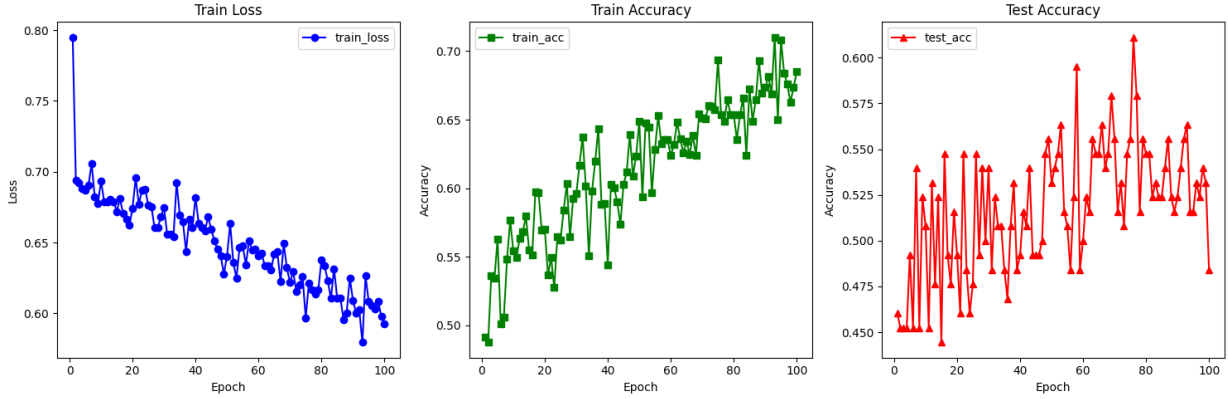

(a) Session 1

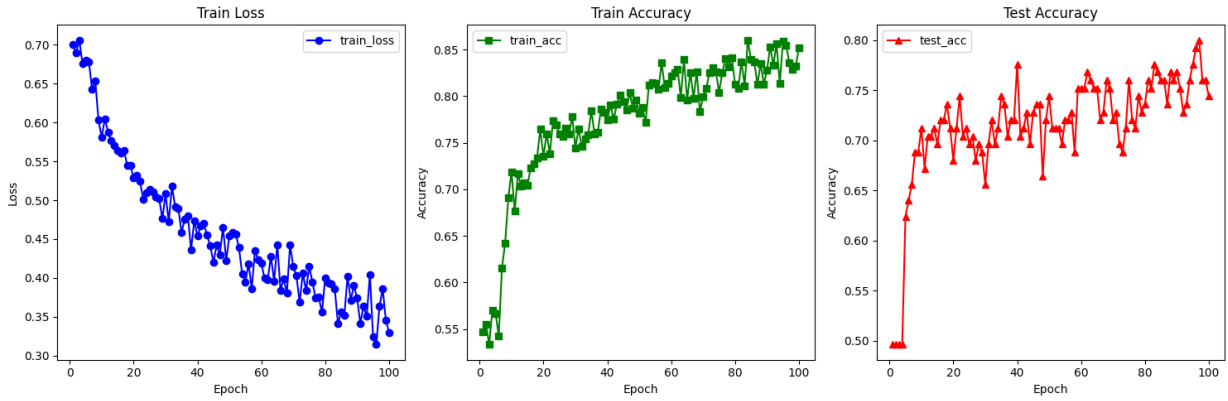

(b) Session 2

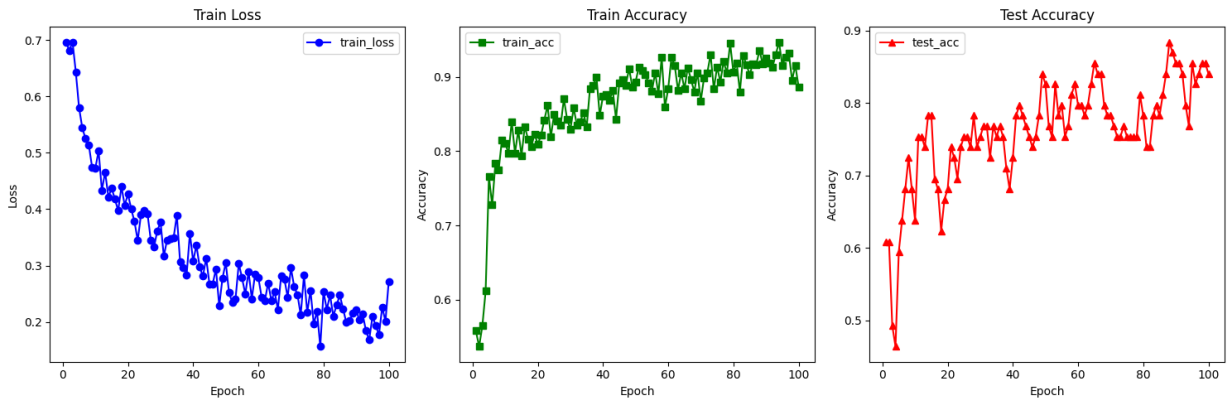

(c) Session 3

Figure 6. Training loss and accuracy curves, along with test accuracy curves (validation), for the three experimental sessions: (a) Session 1, (b) Session 2, and (c) Session 3.

3.4 Hardware and Platforms

All experiments were conducted on a workstation located in Hung Lab, Bainer Hall, at the University of California, Davis. The workstation was equipped with two NVIDIA GeForce RTX 4090 GPUs (24 GB VRAM each), an Intel Core i9-14900KS CPU, and 128 GB of RAM. The software environment ran Ubuntu 22.04 with CUDA 12.1. The main scripts were implemented in Python, and all deep learning models were trained and deployed using PyTorch 2.1.0.

4. Results and Discussion

4.1 Zero-shot Foundation Model Analysis

We estimated the quality of zero-shot fish ROI extraction via foundation model based on human intuitively as shown in Figure 7. Basically, it keeps the fish main phenotype and morphology information including main fish body and fish fins. Some uncomplete margin shape segmented of fish like the caudal fin will be eliminated by the subsequently data augmentation like random erasing.

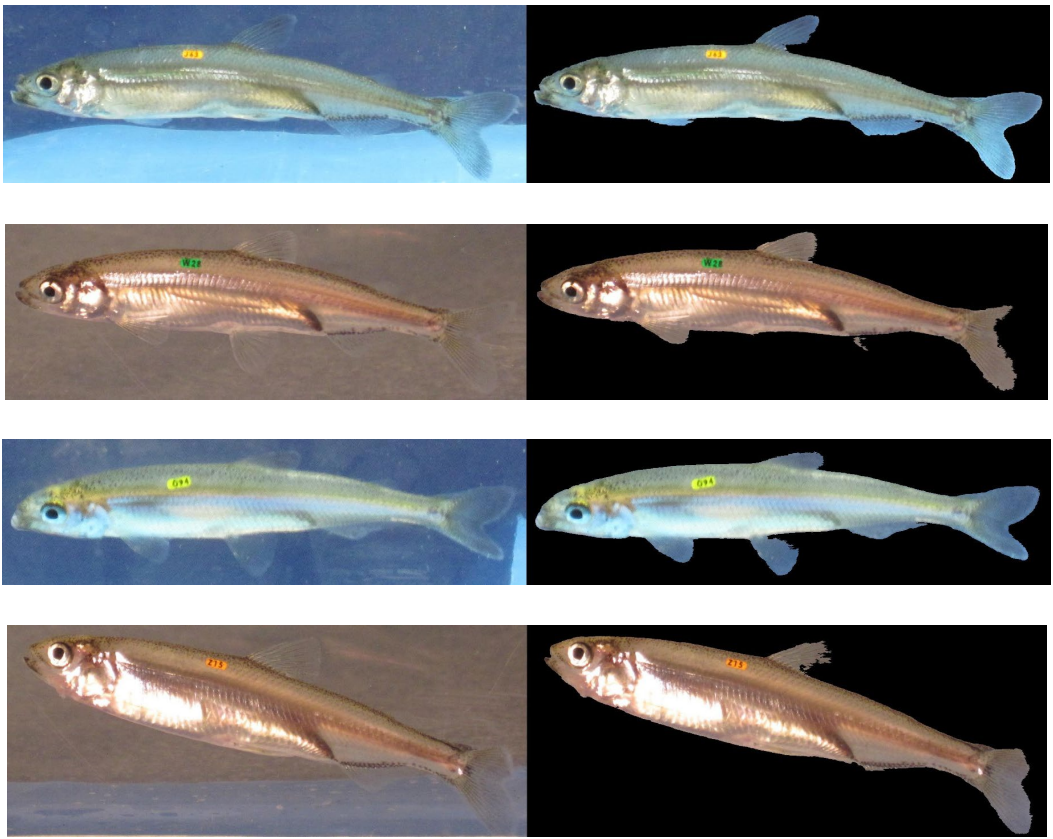

Figure 7. ROI extraction results from the ROI extraction module. The left column represents the original images, and the right column represents the corresponding fish ROIs.

The reason we perform fish ROI extraction is that we found AI driven decision making tends to focus on background bias, such as differences in tank bottom appearance, while neglecting the morphological differences between female and male fish, as shown in Figure 8. We finetune last several layers of ResNet for sex identification of delta smelt.

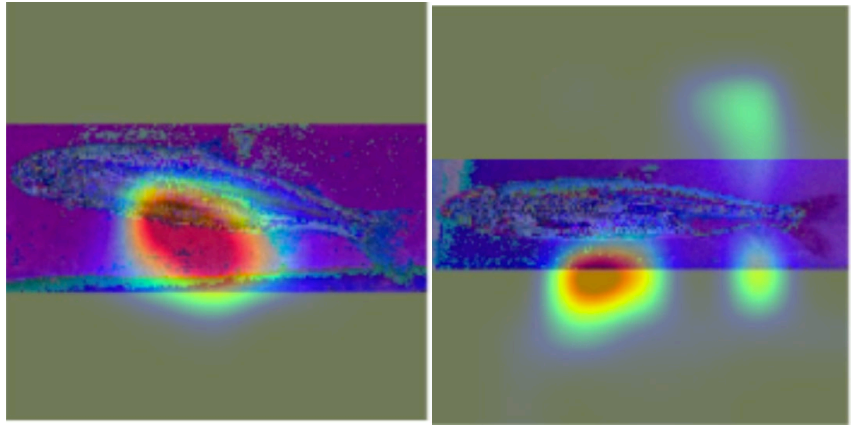

Figure 8. Class activation mapping for fish sex identification by finetuning of the last fully connected layer (left) and the last convolutional layers of the ResNet 34 model (right)

4.2 Prototype-based Interpretability Analysis

FishProtoNet was applied to previously unseen the fish images in the FCCL delta smelt sex test dataset. The decision-making process for fish sex identification was interpreted using the 3 layers depth neural tree model, as illustrated in Figure 9. FishProtoNet identifies the fish sex by comparing the features extracted from the input image with learned prototypes at each node of the tree model. Each node in the neural tree corresponds to a learned prototype that captures salient fish features relate to distinguish female and males of fish, as shown in Figure 8. Each path from root node to leaf node represents a step-by-step decision-making chain based on feature templates matching. At the leaf level, the model outputs predicted probabilities for fish sex (female or male). The final fish sex prediction is determined by the class with the highest aggregated probability, obtained by summing the probabilities across the corresponding leaf nodes.

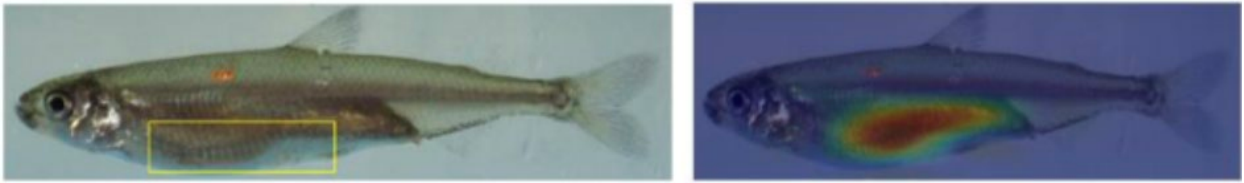

Figure 9. Visualization of a learned prototype. Left: the prototype region in the original image marked by a bounding box; Right: the corresponding activation heatmap of the prototype.

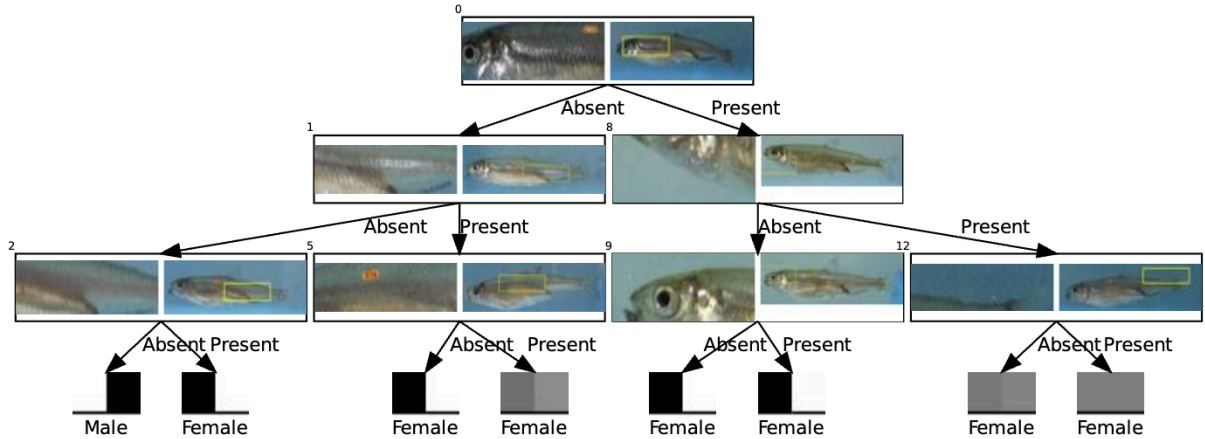

Figure 10. Interpretable visualization of the FishProtoNet decision-making chain during inference for delta smelt sex identification on a test image. Each internal node displays a learned prototype (left) and its corresponding source image from the training dataset. "Present" indicates that the similarity between the prototype and the test image exceeds a predefined threshold, whereas "Absent" indicates the opposite.

4.3 Data Augment Analysis

Effective feature extraction from fish ROIs provides essential patterns for accurate sex identification. However, some zero-shot generated fish ROIs were unsatisfactory, particularly in cases where fish body boundaries were ambiguous, which may negatively affect fish sex identification. To address this uncertainty in ROI extraction, data augmentation like random erasing were employed to enhance model robustness. In addition, random erasing helps to reduce the potential influence or bias introduced by VIA tags or artifacts in the dataset.

Different data augmentation strategies were evaluated and compared on the FCCL delta smelt sex dataset Session 3. ResNet-50 was fine-tuned on the FCCL delta smelt sex training dataset using the corresponding data augmentation strategies. All augmentation experiments were conducted using an initial learning rate of 0.001, a batch size of 16 and the Adam optimizer. OneCycleLR learning rate scheduler was adopted over 50 training epochs. As shown in Table 1, the combination of random rotation, horizontal flip, random crop, and random erasing, together with the base augmentation (Image resizing, normalization, and padding) was found to achieve the highest accuracy in delta smelt sex classification.

Table 1. Comparison of model performance under different data augmentation settings on the FCCL delta smelt sex dataset (Session 3).

| Experiment ID | Augmentation | Model | Accuracy (%) | F1 scores (%) |
| --- | --- | --- | --- | --- |
| 0 (Base) | Resize, Normalization and Padding | ResNet-50 | 71.01 | 69.14 |
| 1 | Base + Random Rotation (15°) and Random Horizontal Flip | ResNet-50 | 76.81 | 75.66 |
| 2 | Base + Color jitter | ResNet-50 | 63.77 | 50.21 |
| 3 | Base + Random Crop | ResNet-50 | 79.71 | 78.40 |
| 4 | Base + Random Erasing | ResNet-50 | 73.91 | 71.25 |
| 5 | Base + Rotation + Random Horizontal Flip + Random Crop + Random Erasing | ResNet-50 | 81.82 | 81.06 |

4.4 Model Comparisons

We compared our proposed model FishProtoNet with baseline models, including ResNet-50 and Vision Transformer (ViT) (Dosovitskiy et al., 2020), for delta smelt sex identification. Specifically, a ResNet-50 model pre-trained on ImageNet and a ViT-Base model pre-trained on ImageNet-21K were fine-tuned on the FCCL delta smelt sex training dataset to predict the delta smelt sex identity. The comparative results are summarized in the Table 2.

Table 2: Comparison of model performance across the three sessions of the Delta Smelt dataset

| Model | Session 1 (%) | | Session 2 (%) | | Session 3 (%) | |
|---|---|---|---|---|---|---|
| | Accuracy | F1 Score | Accuracy | F1 Score | Accuracy | F1 Score |
| ResNet-50 | 49.21 | 48.88 | 54.40 | 52.34 | 81.82 | 81.06 |
| ViT | 51.50 | 50.46 | 81.60 | 81.60 | 85.50 | 84.57 |
| FishProtoNet | 54.76 | 54.62 | 74.40 | 74.27 | 81.16 | 79.43 |

As the Table 2 illustrated, FishProtoNet achieved classification accuracies of 54.76%, 74.40%, and 81.16% for Session 1,2, and 3, respectively, with corresponding F1 scores of 54.62, 74.27, and 79.43. These results indicates that the FishProtoNet is effective in identifying the sex of delta smelt during the early spawning (Session 2) and post-spawning stages (Session 3), but less effective during the subadult stage (Session 1). A possible explanation is that phenotypic or morphological differences become more pronounced as the fish mature and gonadal development progresses.

FishProtoNet outperformed ResNet-50 in both accuracy and F1-score across all three sessions. However, compared with ViT, FishProtoNet exhibited lower performance in Sessions 2 and 3, while providing greater transparency in decision-making for fish sex prediction. This performance gap may be attributed to the use of ResNet-50 as the feature extraction backbone in FishProtoNet, which has limited representational capacity than ViT model. ViT models benefit from a larger number of parameters and pre-training on large-scale dataset, enabling stronger feature representation compared with convolutional neural networks. In addition, transformers capture global contextual information through attention mechanisms, whereas convolutional neural networks rely on local receptive fields. In future work, integrating a ViT backbone into FishProtoNet as the feature extraction module may further improve performance of fish sex identification.

4.5 Limitations and Future Work

Despite these improvements achieved by FishProtoNet, accurately identifying the sex of delta smelt at the subadult stage (Session 1) remains a persistent challenge. This limitation is likely due to less pronounced morphological differences between male and female fish at early developmental stages. Incorporating multi-modal imaging data, such as infrared or ultrasound images, may help capture additional physiological cues and further enhance fish sex identification accuracy for individuals at this stage.

5. Conclusion

In this study, we proposed FishProtoNet, an interpretable prototype-based AI framework that integrates visual foundation models for robust and non-invasive fish sex identification. By combining grounding DINO and SAM2, the framework enables the extraction of high-quality fish ROIs. Optimal data augmentation strategies were systematically explored to improve the model robustness. Experiments results show that FishProtoNet achieved 74.40% accuracy and a 74.27% F1-score in the early spawning stage, increasing to 81.16% accuracy and 79.43% F1-score in the post-spawning stage. However, the model exhibited limited performance for subadult delta smelt, indicating that morphological differences at early development stage may not be sufficiently distinctive for accurate delta smelt sex identification using the current approach. Future work will explore the integration of multimodal inputs, such as near-infrared imaging, as well as the incorporating of additional temporal training data across multiple life stages to enhance fish sex identification at earlier life stages.

Credit Authorship Contribution Statement

ZM: Conceptualization, Methodology, Software Development, Experiments, Results Analysis, Writing – original draft, Writing – review & editing. TCH: Supervision, Conceptualization, Method Validation, Writing – review & editing, Resource Provision, Project Administration.

Declaration of Competing Interest

The authors declare that they have no known competing financial interests or personal relationships that could have appeared to influence the work reported in this paper.

Acknowledgements

We would like to express our sincere gratitude to the researchers at the Fish Conservation and Culture Laboratory (FCCL), University of California, Davis, for providing the delta smelt dataset and supporting the data collection process. We are also deeply grateful to Professor Zhe Zhao from the Department of Computer Science, University of California, Davis, for his valuable suggestions and insightful feedback on this manuscript. Additionally, we thank PhD student Payam Farhadi, Lena Nguyen, and Professor Iman Soltani at UC Davis for their support throughout the course of this work. The production of delta smelt was supported by the U. S. Bureau of Reclamation (No. R20AC00027).